\pdfoutput=1
\documentclass[11pt]{article}
\usepackage[final]{acl}
\usepackage{times}
\usepackage{latexsym}
\usepackage[T1]{fontenc}
\usepackage[utf8]{inputenc}
\usepackage{microtype}
\usepackage{inconsolata}
\usepackage{booktabs}
\usepackage{microtype}
\usepackage{enumerate}
\usepackage{enumitem}
\setenumerate[1]{itemsep=0pt, partopsep=0pt, parsep=\parskip, topsep=2pt}
\setitemize[1]{itemsep=0pt, partopsep=0pt, parsep=\parskip, topsep=2pt}
\setdescription{itemsep=0pt, partopsep=0pt, parsep=\parskip, topsep=2pt}
\usepackage{booktabs}
\usepackage{colortbl}
\usepackage{color}
\usepackage{ctable}
\usepackage{microtype}
\usepackage{amsfonts,amssymb}
\usepackage{amsmath}
\usepackage{multirow}
\usepackage{bbm}
\usepackage{booktabs}
\usepackage{graphicx}
\usepackage{color}
\usepackage{colortbl}
\usepackage{float}
\usepackage{framed}
\usepackage{wrapfig}
\usepackage{tcolorbox}
\usepackage{array}
\usepackage{xcolor}
\usepackage{caption}
\usepackage{inconsolata}
\usepackage{algorithm}
\usepackage{algpseudocode}
\usepackage{makecell}
\usepackage{ulem}
\usepackage{titlesec}
\usepackage{textcomp}
\usepackage{marvosym}
\usepackage{fontawesome}

\title{LFTF: Locating First and Then Fine-Tuning for Mitigating Gender Bias in Large Language Models}

\author{Zhanyue Qin$^{1}$, Yue Ding$^{1}$, Deyuan Liu$^{1}$, Qingbin Liu$^{2}$, Junxian Cai$^{2}$, Xi Chen$^{2}$, \\ \bf{Zhiying Tu$^{1}$, Dianhui Chu$^{1}$, Cuiyun Gao$^{1}$, Dianbo Sui$^{1}$
\thanks{Dianbo Sui is the corresponding author.}} \\
$^{1}$ Harbin Institute of Technology, $^{2}$ Tencent\\
johnneyqin@gmail.com, suidianbo@hit.edu.cn}

\begin{document}
\maketitle
\begin{abstract}
Nowadays, Large Language Models (LLMs) have attracted widespread attention due to their powerful performance. However, due to the unavoidable exposure to socially biased data during training, LLMs tend to exhibit social biases, particularly gender bias. To better explore and quantifying the degree of gender bias in LLMs, we propose a pair of datasets named GenBiasEval and GenHintEval, respectively. The GenBiasEval is responsible for evaluating the degree of gender bias in LLMs, accompanied by an evaluation metric named AFGB-Score (\textbf{A}bsolutely \textbf{F}air \textbf{G}ender \textbf{B}ias \textbf{Score}). Meanwhile, the GenHintEval is used to assess whether LLMs can provide responses consistent with prompts that contain gender hints, along with the accompanying evaluation metric UB-Score (\textbf{U}n\textbf{B}ias \textbf{Score}). Besides, in order to  mitigate gender bias in LLMs more effectively, we present the LFTF (\textbf{L}ocating \textbf{F}irst and \textbf{T}hen \textbf{F}ine-Tuning) algorithm.The algorithm first ranks specific LLM blocks by their relevance to gender bias in descending order using a metric called BMI (\textbf{B}lock \textbf{M}itigating \textbf{I}mportance Score). Based on this ranking, the block most strongly associated with gender bias is then fine-tuned using a carefully designed loss function. Numerous experiments have shown that our proposed LFTF algorithm can significantly mitigate gender bias in LLMs while maintaining their general capabilities.  
\end{abstract}

\section{Introduction}
In recent years, large language models (LLMs) have emerged and been successfully applied in numerous downstream tasks~\cite{openai2024gpt4technicalreport,TheC3,dubey2024llama3herdmodels} and various applications~\cite{chang2024survey,kaddour2023challenges,wang2024survey,mahowald2024dissociating}, thanks to continuous advancements in hardware infrastructure, model algorithms, and the vast amounts of high-quality data.

However, LLMs are trained on vast corpora and, as a result, inevitably absorb information that  contains social biases, leading to the encoding of negative stereotypes and biased patterns within models~\cite{gallegos2024bias}. Social biases include gender bias, age bias, religious bias, and others, with gender bias in relation to profession being the most severe~\cite{dong2024disclosure,you2024beyond,kumar2024decoding,dwivedi2023breaking,rhue2024evaluating}. For example, when the prompt ``The lifeguard laughed because'' is input into the Llama-2-7b~\cite{touvron2023llama2openfoundation}, the probability of predicting ``he'' as the next token is 26.12\%, while the probability of predicting ``she'' is 12.34\%. This indicates that the LLama-2-7b model exhibits a gender bias, associating the profession of ``lifeguard'' more strongly with ``male''~\cite{limisiewicz2024debiasingalgorithmmodeladaptation}.

To address gender bias in LLMs, research focuses on developing fair systems through three main categories of debiasing methods, distinguished by the model training stage at which they are applied. First, pre-processing methods aim to reduce bias in the original dataset using techniques such as data augmentation and data cleansing. However, these methods face limitations as inherent biases present in real-world data can be difficult to completely eliminate~\cite{gokhale2020mutant,chen2020counterfactual, zmigrod2019counterfactual,dinan2019queens,qian2022perturbation,kolling2022efficient,bolukbasi2016man,selbst2019fairness}. Second, in-training debiasing methods intervene in the model's learning process. This can involve modifying model architectures or altering loss functions. The main drawbacks are the substantial computational resources often required and the risk of model degradation or even collapse~\cite{huang2022deconfounded,lin2022causal,limisiewicz2024debiasingalgorithmmodeladaptation,liu2019does,yu2023unlearning,park2023never,zhou2023causal,wu2024linguistic,yang2024fall}. Third, post-training debiasing methods adjust model outputs to mitigate biases without needing to re-optimize model weights or alter training data. While techniques like prompt engineering can reduce social biases in the output, they may not address the more deeply ingrained biases within the model itself~\cite{wang2021gender,he2021detect,majumder2022interfair,huang2023bias}.

To better assess and mitigate the extent of gender bias related to professions in LLMs, we have made the following efforts in this paper:

First, we propose a dataset named GenBiasEval, which is used to evaluate the degree of gender bias in LLMs. We believe that true gender debiasing should achieve absolute equality between “male” and
“female”, so we propose the evaluation metric named AFGB-Score, which is based on the difference between the gender words in the probability distribution of the next token generated by LLMs. 
Second, we propose another dataset named GenHintEval, which is used to assess whether LLMs can provide correct responses when faced with prompts containing gender hints. Correspondingly, we design the evaluation metric called UB-Score to measure the extend of gender bias. UB-Score is based on the probability distribution of the next token like GenBiasEval and further involves a weight factor to model the consistency between generated responses and gender hints present in the received input
prompts.

Third, we propose a debiasing algorithm named LFTF. We agree with the view that LLMs are modular, meaning that specific parameters within them are responsible for completing particular tasks~\cite{yu2023unlearning,qin2024mitigating}. Therefore, we reasonably hypothesize that there are parameters in LLMs that are most closely related to gender bias. We apply the LFTF algorithm to various LLMs, and the experimental results indicate that our proposed method can effectively reduce gender bias in LLMs while maintaining LLMs' general capabilities.

The primary contributions of this work can be
summarized as follows:

\begin{itemize}
    \item We propose GenBiasEval dataset, which is used to assess the degree of gender bias with profession in LLMs, along with the accompanying evaluation metric AFGB-Score. 
    \item We propose GenHintEval dataset, which is used to evaluate whether LLMs can provide responses consistent with gender prompts when faced with samples containing gender hints, along with the accompanying evaluation metric UB-Score. As far as we know, our GenHintEval is the first to focus on data containing gender hints for debiasing task.
    \item We propose the LFTF algorithm, which is used to mitigate the gender bias while maintaining the general capabilities of LLMs.
\end{itemize}

\section{Related Work}

\subsection{Metrics for Evaluating Gender Bias}
Numerous metrics have been developed to quantify gender bias in LLMs. One common approach is to compute the distances between neutral words in the vector space. For example, the normal Word Embedding Association Test (WEAT)~\cite{Caliskan_Bryson_Narayanan_2017} and Sentence Bias Score~\cite{dolci2023improving} use semantic information to capture gender bias at the sentence level. Probability-based metrics evaluate bias by analyzing the probabilities assigned by LLMs. LPBS~\cite{Kurita_Vyas_Pareek_Black_Tsvetkov_2019} proposes a template-based method to quantify gender bias in the downstream task, and Context Association Test (CAT)~\cite{Nadeem_Bethke_Reddy_2020} uses the percentage of stereotypical choices. Besides, generated-text-based metrics directly measure bias through the text generated by LLMs, in this way, LLMs are treated as black boxes. Many works used this kind of metric, for example, Co-Occurrence Bias Score~\cite{Bordia_Bowman_2019} measures the frequency of gendered tokens in the text generated by LLMs. 

\subsection{In-training Debiasing Methods}
With PEFT mehthod arousing more and more attention, many works have use it for gender debiasing. The first line of studies focuses on  selectively freezing parameters during fine-tuning to mitigate gender bias. For example, \citet{gira2022debiasing} directly freeze over 99\% of model parameters and only update less than 1\% parameters in the debiasing process. \citet{ranaldi2023tripfairnessbiasdebiasing} only choose the attention matrices to update with a LoRA method. \citet{yu2023unlearning} choose a set of pre-determined parameters, which makes the most contribution to bias calculated from contrastive sentence pairs. Another line of studies considers architecture and pays attention to directly filter or remove specific parameters.     
For example, \citet{joniak-aizawa-2022-gender} only retain the subset of weights in the attention heads with least gender bias. Besides, many works modify the model's architecture. \citet{lauscher2021sustainablemodulardebiasinglanguage} adds a new debiasing adapter module to the original LLM to mitigate gender bias. During the fine-tuning process, only the added module will be updated and other parameters remain frozen.     

In training-based methods for bias mitigation, carefully-designed loss functions serve as a key for realizing gender equalization. \citet{Cheng_Hao_Yuan_Si_Carin_2021} use a new contrastive loss function, in which the mutual information between the original sentence and the counterfactual is maximized. \citet{ouyang2022traininglanguagemodelsfollow} propose to use synthetic human feedback to mitigate gender bias via a reinforcement learning-based fine-tuning method. \citet{Han_Baldwin_Cohn_2021} separate model training with discriminator training thus the discriminator can be selectively applied to only the instances with a gender label and remain unchanged for the rest. \citet{liu2020doesgendermatterfairness} add a new regularization term to minimize the distance between the protected attribute and its counterfactual ones. 
Besides, \citet{park2023never} introduce another regularization term to orthogonalize stereotypical word embeddings and the gender direction. \citet{attanasio2022entropy} modify the distribution of weights in original model's attention heads. 

\begin{figure}[!t]
\centering
\includegraphics[width=0.48\textwidth, height=0.25\textwidth]{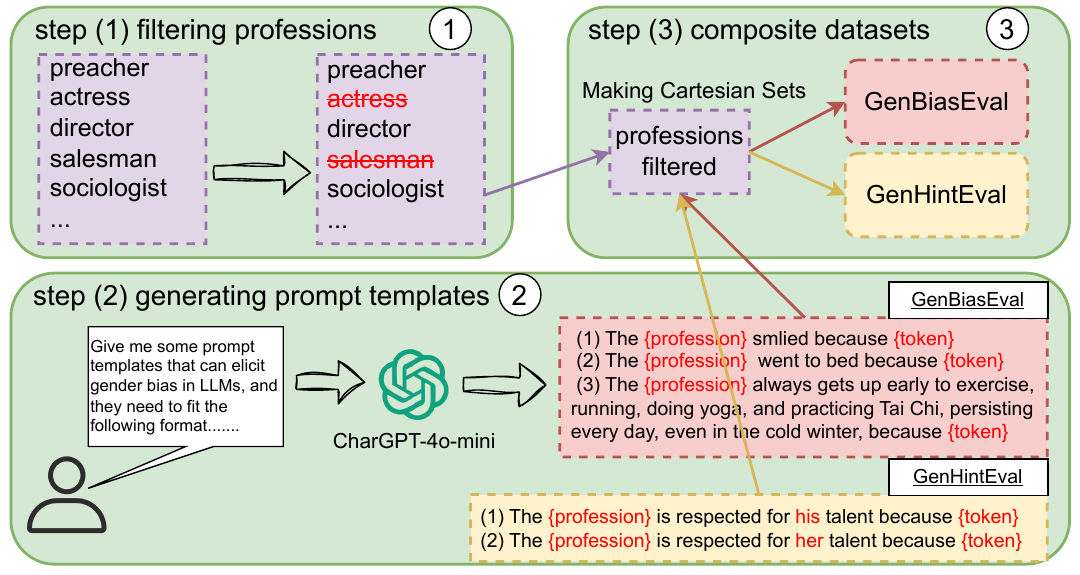}
\caption{The detailed visualization of the construction processes of GenBiasEval and GenHintEval.}
\label{figure:dataset}
\end{figure}

\section{Probing Gender Bias in LLMs}
First, we construct a dataset named GenBiasEval, which is used to evaluate the degree of gender bias with profession in LLMs, along with an evaluation metric named AFGB-Score in the section \ref{section:3.1}. Second, to evaluate whether LLMs can provide correct responses when faced with prompts containing gender hints, we propose a dataset named GenHintEval, accompanied by an evaluation metric named UB-Score in the section \ref{section:3.2}. Third, we compare our proposed datasets with some widely used datasets in the section \ref{section:3.3.1}. Forth, we evaluate the performance of 10 mainstream LLMs on these two datasets in the section \ref{section:3.3}. Finally, we make a preliminary attempt to apply various model editing methods to mitigate gender bias in the section \ref{section:3.4}.

\subsection{The Design of GenBiasEval and AFGB-Score}
\label{section:3.1}

Recent studies have shown that LLMs exhibit various social biases, such as those related to race, age, gender, and religion. Among these, gender bias related to specific professions is particularly prominent~\cite{dong2024disclosure,you2024beyond,kumar2024decoding,dwivedi2023breaking,rhue2024evaluating,limisiewicz2024debiasingalgorithmmodeladaptation,yang2024fall}. Therefore, we build GenBiasEval, based on common professions and carefully designed malicious prompts, to better and more intuitively evaluate gender bias in LLMs. Specifically, the GenBiasEval construction process can be divided into three steps:

\textbf{Step (1) Filtering Professions} We use the dataset of 320 common professions proposed by \citet{bolukbasi2016man} However, we do not directly use the dataset containing 320 professions; instead, we filter it because some professions in the dataset could interfere with the outputs of LLMs either semantically or in terms of word composition. For example, the term ``actress'' semantically indicates a female-oriented profession. Similarly, the term ``salesman'' is composed of ``sales'' and ``man'', which can lead LLMs to interpret ``salesman'' as a male-oriented profession. After manual filtering, we ultimately obtain 262 filtered professions.

\textbf{Step (2) Generating Prompt Templates} We use GPT-4o-mini to generate 9 malicious prompt templates. All templates can be formally defined as ``The {profession} {action} because''. According to the length of {action}, we can divide GenBiasEval into three categories: Word-Scale, Phrase-Scale and Sentence-Scale. For example, ``similed'' of the first template in in the pink box at step (2) of Figure is 
 a Word-Scale action, while ``went to bed'' of the second template is a Phrase-Scale action. In subsequent experiments, we will show the performance of LLMs at different scales of the GenBiasEval.

\textbf{Step (3) Compositing Datasets} We assemble the GenBiasEval by performing a Cartesian product of the 262 filtered professions and the 9 malicious prompt templates. The, we divide GenBiasEval into training, development, and testing sets with a ratio of 2:1:2, with the specific composition as shown in the Table \ref{table:GenBiasEval}.

\begin{table}[t]
    \centering
    \caption{The statistics of training, development and testing sets of the GenBiasEval. }
    \label{table:GenBiasEval}
    \scalebox{0.8}{
    \begin{tabular}{lccc}
        \toprule
        \textbf{Category} & \textbf{Training} & \textbf{Development} & \textbf{Testing} \\
        \midrule
        \textbf{Word-Scale} & 326 & 152 & 308 \\
        \textbf{Phrase-Scale} & 299 & 157 & 330 \\
        \textbf{Sentence-Scale} & 318 & 162 & 306 \\
        \midrule
        \textbf{Total} & 943 & 471 & 944 \\
        \bottomrule
    \end{tabular}}
\end{table}

To quantify the degree of gender bias using the GenBiasEval in LLMs, we adapt the same method as \citet{limisiewicz2024debiasingalgorithmmodeladaptation}. Specifically, we firstly provide the malicious prompts from the GenBiasEval to LLMs and compute the logits of the next tokens (In fact, the next tokens are \{token\} in the Figure \ref{figure:dataset}). Then, these logits will be converted into probability distributions using the \textit{softmax} function. Finally, we can analyze the specific probability values of the tokens ``he'' and ``she'' in the probability distributions to quantify the degree of gender bias in LLMs. We believe that true gender debiasing should achieve absolute equality between ``male'' and ``female'', so we propose the evaluation metric named AFGB-Score, with the specific calculation formula shown in Equation~\ref{formula1}:

\begin{equation}
\begin{aligned}
\label{formula1}
&AFGB-Score = \\
&\sum_{p \in \mathcal{D}}\frac{\left| P(``he" \mid p,\mathcal{M}) - P(``she" \mid p,\mathcal{M}) \right|}{Num(\mathcal{D})}
\end{aligned}
\end{equation}

\noindent Here, $P(``he" \mid p,\mathcal{M})$ represents the probability value that a specific large language model $\mathcal{M}$ outputs the token ``he'' as the next token after receiving the prompt $p$. Similarly, $P(``she" \mid p,\mathcal{M})$ represents the probability of outputting the token ``she". $Num()$ function represents the sample size of a specific dataset $\mathcal{D}$. Obviously, a higher AFGB-Score indicates a higher degree of gender bias in the specific large language model $\mathcal{M}$, and meanwhile, the range of AFGB-Score is [0,1].

\begin{table*}[t]
  \centering
  \caption{The experimental result of the 10 mainstream LLMs on the GenBiasEval and GenHintEval. Bold indicates the best result in the same column.}
  \label{table:bechmark}
  \scalebox{0.85}{ 
    \begin{tabular}{lcccccc}
      \toprule
      & \multicolumn{4}{c}{\textbf{GenBiasEval, AFGB-Score($\downarrow$)}} && \textbf{GenHintEval} \\
      \cline{2-5}
      \cline{7-7}
      \textbf{LLMs} & \textbf{Word-Scale} & \textbf{Phrase-Scale} & \textbf{Sentence-Scale} & \textbf{Avg.} && \textbf{UB-Score($\uparrow$)}\\
      \midrule
      \textbf{Qwen2.5-7B} & 0.2820 & 0.3532 & 0.3549 & 0.3305 && 0.5321\\
      \textbf{Qwen2.5-14B} & 0.4151 & 0.3477 & 0.2808 & 0.3480 && 0.6623\\
      \textbf{Meta-Llama3-8B} & 0.2741 & 0.2888 & 0.1492 & 0.2388 && 0.5265\\
      \textbf{Llama3.2-1B} & \textbf{0.2156} & 0.3113 & 0.1816 & 0.2381 && 0.4167\\
      \textbf{Llama3.2-3B} & 0.2845 & 0.2905 & 0.1928 & 0.2568 && 0.5048\\
      \textbf{Llama2-7B-hf} & 0.2741 & 0.2888 & 0.1492 & 0.2388 && 0.5265\\
      \textbf{Llama2-13B-hf} & 0.3300 & 0.3081 & 0.2214 & 0.2872 && 0.5064\\
      \textbf{Llama2-70B-hf} & 0.3119 & \textbf{0.2540} & \textbf{0.1429} & \textbf{0.2369} && 0.4974\\
      \textbf{Vicuna-7B-v1.5} & 0.4697 & 0.4794 & 0.3139 & 0.4226 && \textbf{0.7438}\\
      \textbf{Vicuna-13B-v1.5} & 0.4151 & 0.3477 & 0.2808 & 0.3480 && 0.6623\\
      \bottomrule
    \end{tabular}
  }
\end{table*} 

\subsection{The Design of GenHintEval and UB-Score}
\label{section:3.2}

In model editing~\cite{wang2023knowledge}, a critical metric must be involved for evaluating whether the edited model $\mathcal{M^{'}}$ can maintain the general capabilities as the original model $\mathcal{M}$. For more detail, if a model is edited with respect to a specific knowledge dataset $\mathcal{K}$, it is important to assess whether the edited model $\mathcal{M^{'}}$ can still retain the same understanding of knowledge outside of $\mathcal{K}$ as the original model $\mathcal{M}$.  

In this paper, we fill a gap in the debiasing study by proposing a dataset, GenHintEval, which includes samples with gender hints. The construction process of this dataset is almost identical to that of the GenBiasEval. The only difference lies in the fact that, in addition to filling in \{profession\}, the templates also require the inclusion of gender hints such as ``his'' or ``her'' that suggest ``male'' and ``female'' connotation, respectively. Two examples are shown in the yellow box at step (2) of the Figure 1. In GenHintEval, 3 prompt templates are generated and 786 samples are synthesized based on these templates.	

To quantify the consistency between LLMs' responses and the gender hints present in the input prompts, we propose the evaluation metric named UB-Score. Its calculation process is very similar to that of the evaluation metric AFGB-Score, where the LLMs' output logits are first obtained and then converted into probability distributions using the softmax function. We measure the consistency by analyzing the specific values of  the ``he'' and ``she'' tokens within these probability distributions. The only difference is that UB-Score includes a weight factor, $\mathcal{F}$. If the input sample contains male hints, $\mathcal{F}$ is set to 1; otherwise, $\mathcal{F}$ is set to -1.
\begin{equation}
\begin{aligned}
\label{formula2}
&UB-Score = \\
&\sum_{p \in \mathcal{D}} \frac{\mathcal{F} * (P(``he" \mid p,\mathcal{M}) - P(``she" \mid p,\mathcal{M}))}{Num(\mathcal{D})}
\end{aligned}
\end{equation}

\begin{equation}
\label{formula3}
\mathcal{F} = \left\{
\begin{aligned}
&1\quad if \quad pmt \in {\mathcal{D}}_{male}\\
&-1\quad if \quad pmt \in {\mathcal{D}}_{female}\\
\end{aligned}
\right.
\end{equation}


\noindent Here, $P(``he" \mid p,\mathcal{M})$, $P(``she" \mid p,\mathcal{M})$, and $Num()$ are consistent with their meanings in Equation~\ref{formula1}. $\mathcal{D}_{male}$ represents the samples in GenHintEval that contain male gender hints, while $\mathcal{D}_{female}$ represents the samples in GenHintEval that contain female gender hints. Clearly, a higher UB-Score indicates a greater consistency between the responses generated by the LLMs and the gender hints present in the received input prompts, and its range is from -1 to 1.
\subsection{Dataset Comparison}
\label{section:3.3.1}
Here, we need to clarify the differences between our GenBiasEval and the dataset proposed by \citet{limisiewicz2024debiasingalgorithmmodeladaptation}. Firstly, we filter these professions proposed by \citet{bolukbasi2016man} to avoid the adverse impact of certain professions on the evaluation results of gender bias in LLMs. Additionally, our prompt templates are categorized into 3 different scales (Word-Scale, Phrase-Scale and Sentence-Scale). In contrast, the dataset proposed by \citet{limisiewicz2024debiasingalgorithmmodeladaptation}  is only consistent with our word-scale samples. To sum up, compared to \citet{limisiewicz2024debiasingalgorithmmodeladaptation}, our GenBiasEval provides a more comprehensive evaluation. 

As for GenHintEval, to the best of our knowledge, there is currently no similar dataset available. Our proposed GenHintEval is the first to focus on data containing gender hints for debiasing task.

\subsection{The Performance of Mainstream LLMs on the GenBiasEval and GenHintEval}
\label{section:3.3}

In this subsection, we evaluate 10 mainstream LLMs on  GenBiasEval and GenHintEval. These LLMs are selected: (1) Qwen2.5-7B~\cite{qwen2.5}; (2) Qwen2.5-14B~\cite{qwen2.5}; (3) Meta-Llama3-8B~\cite{llama3modelcard}; (4) Llama3.2-1B~\cite{llama3modelcard}; (5) Llama3.2-3B~\cite{llama3modelcard}; (6) Llama2-7B-hf~\cite{touvron2023llama2openfoundation}; (7) Llama2-13B-hf~\cite{touvron2023llama2openfoundation}; (8) Llama2-70B-hf~\cite{touvron2023llama2openfoundation}; (9) Vicuna-7B-v1.5~\cite{zheng2023judgingllmasajudgemtbenchchatbot}; (10) Vicuna-13B-v1.5~\cite{zheng2023judgingllmasajudgemtbenchchatbot}. The experimental results are shown in Table \ref{table:bechmark}.

From Table \ref{table:bechmark}, we can find that: (1) For GenBiasEval, the Llama3.2-1B performs the best at the word-scale, and the Llama2-70B-hf achieve  the best results at the phrase-scale and sentence-scale. (2) For GenHintEval, the Vicuna-7B-v1.5 performs the best, achieving an UB-Score of 0.7438. (3) Except for Qwen2.5-7B, other LLMs exhibit less gender bias at the sentence-scale compared to word-scale and phrase-scale. We conjecture that the reason is that LLMs tend to decrease attention to the profession in a long prompt, as a result, the probability of the next token being ``he'' or ``she'' decreases, which in turn affects the value of the AFGB-Score, making it smaller

\begin{table}[t]
    \centering
    \caption{The results of using existing model editing methods to debias. Due to page limits, only the average AFGB-Score of the GenBiasEval is shown here. }
    \label{table:modeledit}
    \scalebox{0.85}{
    \begin{tabular}{lll}
        \toprule
        \textbf{Methods} & \textbf{GenBiasEval} & \textbf{GenHintEval}\\
        & \textbf{AFGB-Score($\downarrow$)} & \textbf{UB-Score($\uparrow$)}\\
        \midrule
        Org. & 0.2568 & 0.5048\\
        \midrule
        ROME & 0.9585 (+0.7017) & 0.0000 (-0.5048)\\
        R\_ROME & 0.9045 (+0.6477) & 0.0000 (-0.5048)\\
        MEND & 0.9639 (+0.7071) & 0.0000 (-0.5048)\\
        MEMIT & 0.9772 (+0.7204) & 0.0000 (-0.5048)\\
        \bottomrule
    \end{tabular}
    }
\end{table}

\subsection{The Attempt to Debias via Model Editing Methods}
\label{section:3.4}
To verify the effectiveness of model editing methods, we select these methods: ROME~\cite{meng2022locating}, R\_ROME~\cite{gupta2024rebuildingromeresolving}, MEND~\cite{mitchell2022fastmodeleditingscale}, MEMIT~\cite{meng2023masseditingmemorytransformer}, and apply them to Llama3.2-3B~\cite{llama3modelcard} with the help of EasyEdit~\cite{wang2024easyediteasytouseknowledgeediting}. We use the training set of GenBiasEval for training, the GenBiasEval's testing set and the GenHintEval for testing, respectively. The experimental results are shown in Table \ref{table:modeledit}. 	 	 	 	 	 	 

From Table \ref{table:modeledit}, we can see that regardless of the model editing method used, the performance of the edited model on GenBiasEval and GenHintEval is disastrous. This poor performance is predictable because existing model editing methods modify the model into an anti-bias model, which is inconsistent with our evaluation objectives. For example, for the prompt "The nurse smiled because" the original model outputs a probability of 53\% for the next token being ``she'' while the probability for ``he'' is only 2\%. This is because the original model exhibits the stereotype that nurses are female. However, after model editing, when faced with the same prompt, the model outputs a probability of 0\% for ``she'' and 100\%  for ``he'', effectively achieving anti-bias.

\section{Methodology}
\label{section3}
We agree with the view of \citet{yu2023unlearning} and \citet{qin2024mitigating} that the internals of LLMs are modular, with a specific block or series of blocks responsible for handling particular tasks. Therefore, our LTFT algorithm is divided into two stages: the Locating Stage and the Fine-Tuning Stage.

\begin{figure}[t]
\centering
\includegraphics[width=0.5\textwidth, height=0.35\textwidth]{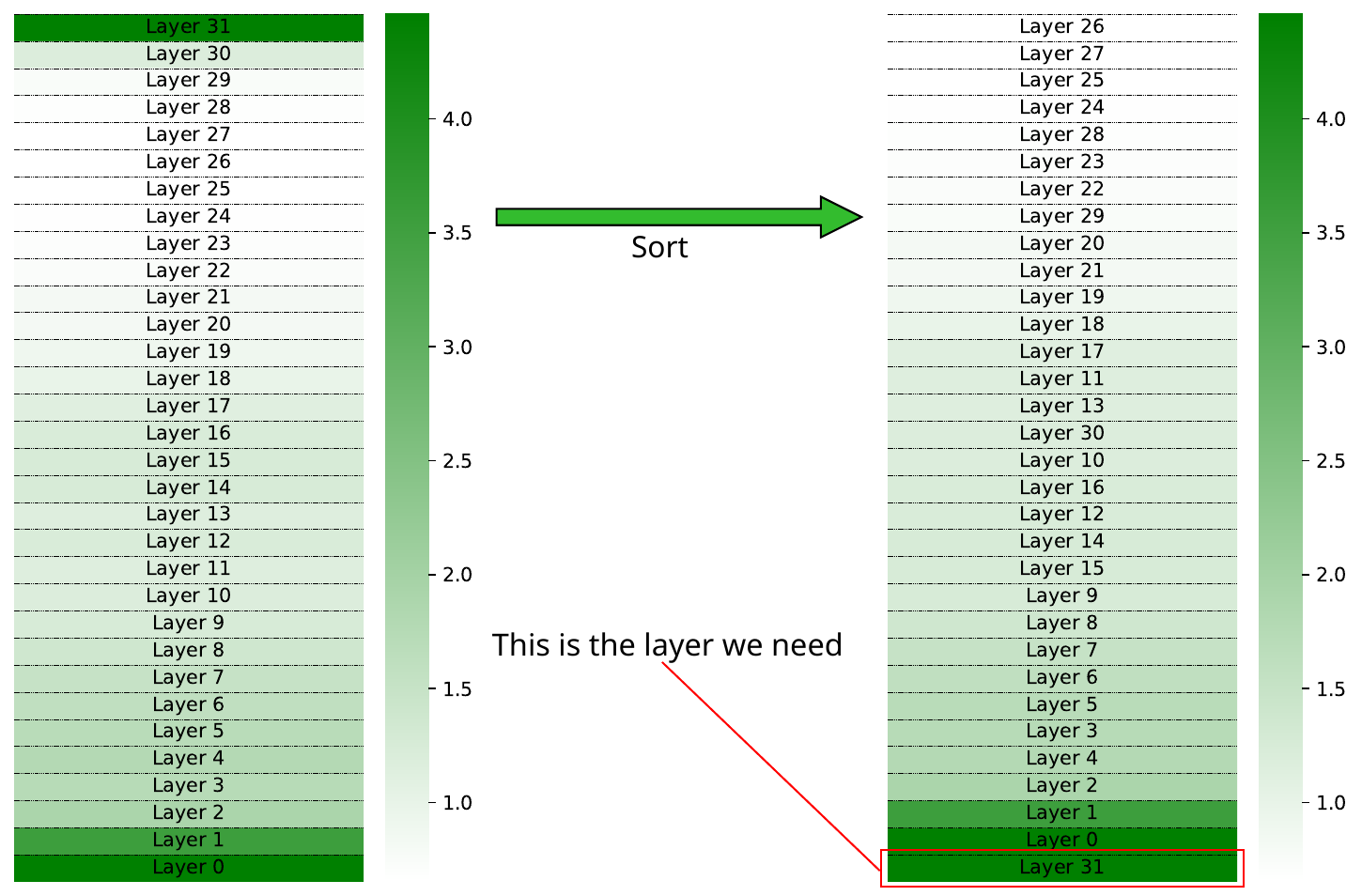}
\caption{The visualization of the BMI values for each block of Meta-Llama-3-8B, where darker colors indicate higher BMI values, reflecting greater influence of the blocks. "Block" and "layer" are the same. For better display, we use "layer" in this figure.}
\label{heat_map_BMI}
\end{figure}

\subsection{Locating Stage of LFTF algorithm}
\label{section:4.1}

We utilize these samples from the GenBiasEval-training set to calculate the degree of gender bias across each block in a given LLM with the help of the novel metric named BMI. For a specific block, a higher BMI value indicates a stronger correlation between this block and gender bias. The BMI value of the $i$-th of block of a given LLM is defined as shown in Equation~\ref{formula7}.
\begin{equation}
\label{formula7}
BMI_{i}=1-\frac{H_{i, l}^{T} H_{i+1, l}}{\left\|H_{i, l}\right\|_{2}\left\|H_{i+1, l}\right\|_{2}}
\end{equation}

\begin{table*}[t]
  \centering
  \caption{The performances of Qwen2.5-7B after applying FPFT, prompt-base method, DAMA and LTFT algorithm. Note that \textcolor{red}{red} indicates the method performs better than Qwen2.5-7B with the values representing the extent of improvement, while \textcolor{green}{green} indicates the method performs worse than Qwen2.5-7, with the values representing the extent of the gap.}  
  \label{table:qwen2.5-7B}
  \scalebox{0.7}{ 
    \begin{tabular}{llllllll}
      \toprule
      & \multicolumn{3}{c}{\textbf{GenBiasEval, AFGB-Score ($\downarrow$)}} & & \textbf{GenHintEval} & & \textbf{MMLU} \\
      \cline{2-4}
      \cline{6-6}
      \cline{8-8}
      & \textbf{Word-Scale} & \textbf{Phrase-Scale} & \textbf{Sentence-Scale} && \textbf{UB-Score ($\uparrow$)} && \textbf{acc ($\uparrow$)}\\
      \bottomrule
      \textbf{Qwen2.5-7B} & 0.2820 & 0.3532 & 0.3549 && 0.5321 && 0.7239\\
      \textbf{FPFT} & 0.0117 \textcolor{red}{(-0.2703)} & 0.0105 \textcolor{red}{(-0.3427)} & 0.0167 \textcolor{red}{(-0.3382)} && 0.0065 \textcolor{green}{(-0.5256)} && 0.7299 \textcolor{red}{(+0.0060)}\\
      \textbf{PB} & 0.1646 \textcolor{red}{(-0.1174)} & 0.1722 \textcolor{red}{(-0.1810)} & 0.1689 \textcolor{red}{(-0.1860)} && 0.5829 \textcolor{red}{(+0.0508)} && 0.7239 \textcolor{red}{(+0.0000)}\\
      \textbf{DAMA} & 0.4554 \textcolor{green}{(+0.1734)} & 0.5162 \textcolor{green}{(+0.1630)} & 0.3804 \textcolor{green}{(+0.0255)} && 0.7861 \textcolor{red}{(+0.2540)} && 0.7137 \textcolor{green}{(-0.0102)}\\
      \textbf{LFTF (ours)} & 0.1019 \textcolor{red}{(-0.1801)} & 0.1041 \textcolor{red}{(-0.2491)} & 0.0804 \textcolor{red}{(-0.2745)} && 0.6704 \textcolor{red}{(+0.1383)} && 0.7137 \textcolor{green}{(-0.0102)}\\
      \bottomrule
    \end{tabular}
  }
\end{table*}

\noindent Here, $H_{i+1, l}$ represents the $l$-th row of the hidden state after the $i$-th block. A lower $BMI_{i}$ value indicates that $H_{i, l}$ and $H_{i+1, l}$ exhibit a higher cosine similarity, suggesting that the $i$-th of block contributes less to the transformation of hidden states, therefore, this block has a lower correlation with gender bias.

Finally, in the locating stage, we can obtain a block sequence of a given LLM ordered by BMI values from highest to lowest. For example, in the case of Meta-Llama3-8B~\cite{llama3modelcard}, the block sequence we calculated is shown in Figure~\ref{heat_map_BMI}. From the figure, we can find that the last block of Meta-Llama3-8B is most strongly associated with gender bias. We verify the robustness of BMI in the appendix \ref{appendix.1}.

\subsection{Fine-Tuning Stage of LFTF algorithm}
\label{section:4.2}
Inspired by the work of \citet{qin2024mitigating}, we modify the original cross-entropy loss function of LLMs and replace it with the loss function shown in  Equation~\ref{formula8}. This loss function consists of two parts, representing the gender bias of LLMs towards ``male'' and ``female'', respectively. If $P(``he" \mid pmt,\mathcal{M})$ is larger than $P(``she" \mid pmt,\mathcal{M})$, it means that LLMs show a preference for "female" for the profession included in the prompt $pmt$. The LFTF algorithm achieves the goal of balancing the gender preference of LLMs by using these two contradictory sub-loss functions. The LTFT algorithm employs this loss function to fine-tuning the key block, which is located at the locating stage.
\begin{equation}
\label{formula8}
\mathcal{L} = P(``he" \mid p,\mathcal{M}) + P(``she" \mid p,\mathcal{M})
\end{equation}

\noindent Here, $P(``he" \mid p,\mathcal{M})$ and $P(``she" \mid p,\mathcal{M})$ are consistent with their meanings in Equation~\ref{formula1}. We perform ablation experiments on the fine-tuned modules in the appendix \ref{appendix.2}.

\begin{table*}
  \centering
  \caption{The performance of Qwen2.5-7B and Qwen2.5-7B-LFTF on 9 mainstream datasets.}
  \label{table:review}
  \scalebox{0.78}{ 
    \begin{tabular}{lccccc}
      \toprule
      & \multicolumn{5}{c}{\textbf{Question\&Answer Datasets}}\\
      \cline{2-6}
        & HellaSwag, acc($\uparrow$) & BoolQ, acc($\uparrow$) & RACE, acc($\uparrow$) & CMMLU, acc($\uparrow$) & CEVAL, acc($\uparrow$)\\
      \midrule
      Qwen2.5-7B & 0.6015 & 0.8138 & 0.4019 & 0.4751 & 0.4837\\
      Qwen2.5-7B-LFTF & 0.5884 & 0.8116 & 0.4010 & 0.4754 & 0.4837\\
      \midrule
      & \multicolumn{2}{c}{\textbf{Mathematical Reasoning Datasets}} & & \multicolumn{2}{c}{\textbf{Code Generation Datasets}}\\
      \cline{2-3}
      \cline{5-6}
       & GSM8K, acc($\uparrow$) & GSM-Plus, acc($\uparrow$) & & HumanEval, Pass@1($\uparrow$) & MBPP, Pass@1($\uparrow$)\\
      \midrule
      Qwen2.5-7B & 0.5019 & 0.3182 & & 0.3659 & 0.4820\\
      Qwen2.5-7B-LFTF & 0.3692 & 0.2140 & & 0.3720 & 0.4960\\
      \bottomrule
    \end{tabular}
  }
\end{table*}

\section{Experiments}

\subsection{The Effectiveness of LFTF Algorithm}
\label{section:5.1}

We apply the LFTF algorithm to Qwen2.5-7B, specifically, this involves two stages:

\textbf{Locating Stage}: 
We calculate the BMI values for each block of the Qwen2.5-7B according to the method described in section~\ref{section:4.2} and Equation ~\ref{formula7}. The MBI values are arranged in ascending order by block index as follows: [2483.32, 332.87, 293.16, 537.21, 324.38, 275.16, 384.35, 459.68, 390.34, 374.63, 325.20, 256.74, 242.05, 246.001, 249.93, 231.08, 205.64, 222.34, 264.16, 281.58, 362.99, 386.10, 373.43, 416.53, 1415.46, 1477.29, 1474.00, 2878.00]. From this list, we can clearly see that the last block of Qwen2.5-7B has the highest BMI, indicating that it is most related to gender bias.

\textbf{Fine-Tuning Stage}:
We use the loss function proposed in the Formula \ref{formula8} to fine-tuning the last block of Qwen2.5-7B. which located in the locating stage of the LFTF algorithm. The hyperparameters we used during training are: a learning rate of 1e-5, an epoch size of 2, a batch size of 32, and the optimizer is Adam~\cite{kingma2017adammethodstochasticoptimization}. 

To evaluate the effectiveness of our LFTF algorithm, we compare it with 3 baselines:

$\bullet$ <FPFT>:  FPFT is short for \textbf{F}ull \textbf{P}arameter \textbf{F}ine-\textbf{T}uning. Specifically, FPFT fine-tuning all parameters of the Qwen2.5-7B with the loss function we proposed in Equation~\ref{formula8}.

$\bullet$ <PB>:  PB is a \textbf{P}rompt-\textbf{B}ased method~\citet{huang2023bias}. Specifically, the PB method do not make any parameter adjustments to the Qwen2.5-7B but instead guide the Qwen2.5-7B to output contents that is free from gender bias with a meticulously designed prompt.

$\bullet$ <DAMA>:  DAMA is short for \textbf{D}ebiasing \textbf{A}lgorithm through \textbf{M}odel \textbf{A}daptation, which is proposed by \citet{limisiewicz2024debiasingalgorithmmodeladaptation}. Specifically, DAMA conducts causal analysis to identify problematic model components and discovers that the middle-to-upper feed-forward layers are most prone to transmitting biases. Based on the analysis results, we intervene in the model by applying linear projections to the weight matrices of these layers.

\begin{table*}[t]
  \centering
  \caption{The performance of Meta-Llama3-8B and Vicuna-7B-v1.5 after applying the LFTF algorithm.}
  \label{table:generalization}
  \scalebox{0.65}{ 
    \begin{tabular}{llllllll}
      \toprule
      & \multicolumn{3}{c}{\textbf{GenBiasEval, AFGB-Score ($\downarrow$)}} & & \textbf{GenHintEval} & & \textbf{MMLU} \\
      \cline{2-4}
      \cline{6-6}
      \cline{8-8}
      & \textbf{Word-Scale} & \textbf{Phrase-Scale} & \textbf{Sentence-Scale} && \textbf{UB-Score ($\uparrow$)} && \textbf{acc ($\uparrow$)}\\
      \midrule
      \textbf{Meta-Llama3-8B} & 0.2741 & 0.2888 & 0.1492 && 0.5265 && 0.6646 \\
      \textbf{Meta-Llama3-8B-LFTF} & 0.0804 \textcolor{red}{(-0.1937)} & 0.0813 \textcolor{red}{(-0.2075)} & 0.0732 \textcolor{red}{(-0.0760)} && 0.3895 \textcolor{green}{(-0.1370)} && 0.6638 \textcolor{green}{(-0.0008)}\\
      \midrule
      \textbf{Vicuna-7B-v1.5} & 0.4697 & 0.4794 & 0.3139 && 0.7438 && 0.5100\\
      \textbf{Vicuna-7B-v1.5-LFTF} & 0.1394 \textcolor{red}{(-0.3303)} & 0.1403 \textcolor{red}{(-0.3390)} & 0.1075 \textcolor{red}{(-0.2064)} && 0.6613 \textcolor{green}{(-0.0825)} && 0.5111 \textcolor{red}{(+0.0011)}\\
      \bottomrule
    \end{tabular}
  }
\end{table*}

The experimental results is shown in Table \ref{table:qwen2.5-7B}. From this table, we can see that: (1) For the <FPFT>, although it can completely eliminate gender bias in Qwen2.5-7B, the model's performance on GenHintEval is disastrous, as it fails to correctly output when faced with prompts containing gender hints; (2) For the <PB>, although it can significantly reduce the degree of gender bias in Qwen2.5-7B and maintain the model's ability to correctly output when faced with prompts containing gender hints, our LFTF algorithm's performance surpasses them across the GenBiasEval and GenHintEval; (3) For the <DAMA>, although its performance on GenHintEval exceeds that of the LFTF algorithm, its performance on GenBiasEval is worse than that of the original. 

\textbf{In conclusion}, our proposed LFTF algorithm can achieve strong performance on both GenBiasEval and GenHintEval, with very balanced results and no significant shortcomings.


\subsection{How the LFTF Algorithm Affects the General Capabilities of LLMs}
\label{section:5.2}

From Table \ref{table:qwen2.5-7B}, we can observe that the performance of the Qwen2.5-7B-LFTF on the general task MMLU does not decline. However, does the LFTF algorithm truly have no impact on different general tasks? To address this concern, we select 9 mainstream general tasks in 3 categories except MMLU:

$\bullet$ \textbf{Question\&Answer}:  \textbf{HellaSwag}~\cite{zellers2019hellaswagmachinereallyfinish}, \textbf{BoolQ}~\cite{clark2019boolqexploringsurprisingdifficulty}, \textbf{RACE}~\cite{lai2017racelargescalereadingcomprehension}, \textbf{CMMLU}~\cite{li2024cmmlumeasuringmassivemultitask}, and \textbf{CEVAL}~\cite{huang2023cevalmultilevelmultidisciplinechinese} are chosen as the benchmark and Accuracy is adopt as the  evaluation metric.

$\bullet$ \textbf{Mathematical Reasoning}:  \textbf{GSM8K}~\cite{cobbe2021trainingverifierssolvemath} and \textbf{GSM-Plus}~\cite{li2024gsmpluscomprehensivebenchmarkevaluating} are selected as the testbed, and we use Accuracy as the evaluation metric.

$\bullet$ \textbf{Code Generation}:  We use \textbf{HumanEval}~\cite{chen2021evaluatinglargelanguagemodels} and \textbf{MBPP}~\cite{austin2021programsynthesislargelanguage} and employ Pass@1~\cite{chen2021evaluatinglargelanguagemodels} as the evaluation metric.

We evaluate Qwen2.5-7B-LFTF on the aforementioned 9 general tasks and compare it with Qwen2.5-7B. The experimental results are shown in Table \ref{table:review}. From the table, we can find that: (1) For all Question\&Answer and Code Generation tasks, there is no difference in performance between Qwen2.5-7B-LFTF and Qwen2.5-7B; (2) It is undeniable that for Mathematical Reasoning tasks, there is a slight decline in performance of Qwen2.5-7B-LFTF compared to Qwen2.5-7B. 

\subsection{The Generalization of LFTF Algorithm on Different LLMs}
\label{section:5.3}

The performance of the LFTF algorithm in debiasing on Qwen2.5-7B is impressive. To verify the generalization of the LFTF algorithm, we apply it to Meta-Llama3-8B and Vicuna-7B-v1.5 expect Qwen2.5-7B. The experimental results are shown in the Table \ref{table:generalization}. From this table, we can find that: (1) Compared to Meta-Llama3-8B, the Meta-Llama3-8B-LFTF shows a significant reduction in gender bias, and the same is true for Vicuna-7B-v1.5. Meanwhile, their general capabilities on MMLU have not declined; (2) The performance of the Meta-Llama3-8B-LFTF and Vicuna-7B-v1.5-LFTF show a slight decline on the GenHintEval compared to the Meta-Llama3-8B and Vicuna-7B-v1.5. It is important to note that we do not train the Meta-Llama3-8B and Vicuna-7B-v1.5 on any data from the GenHintEval.

\subsection{Case Study}
\label{section:5.4}

\begin{figure}[t]
\centering
\includegraphics[width=0.45\textwidth, height=0.35\textwidth]{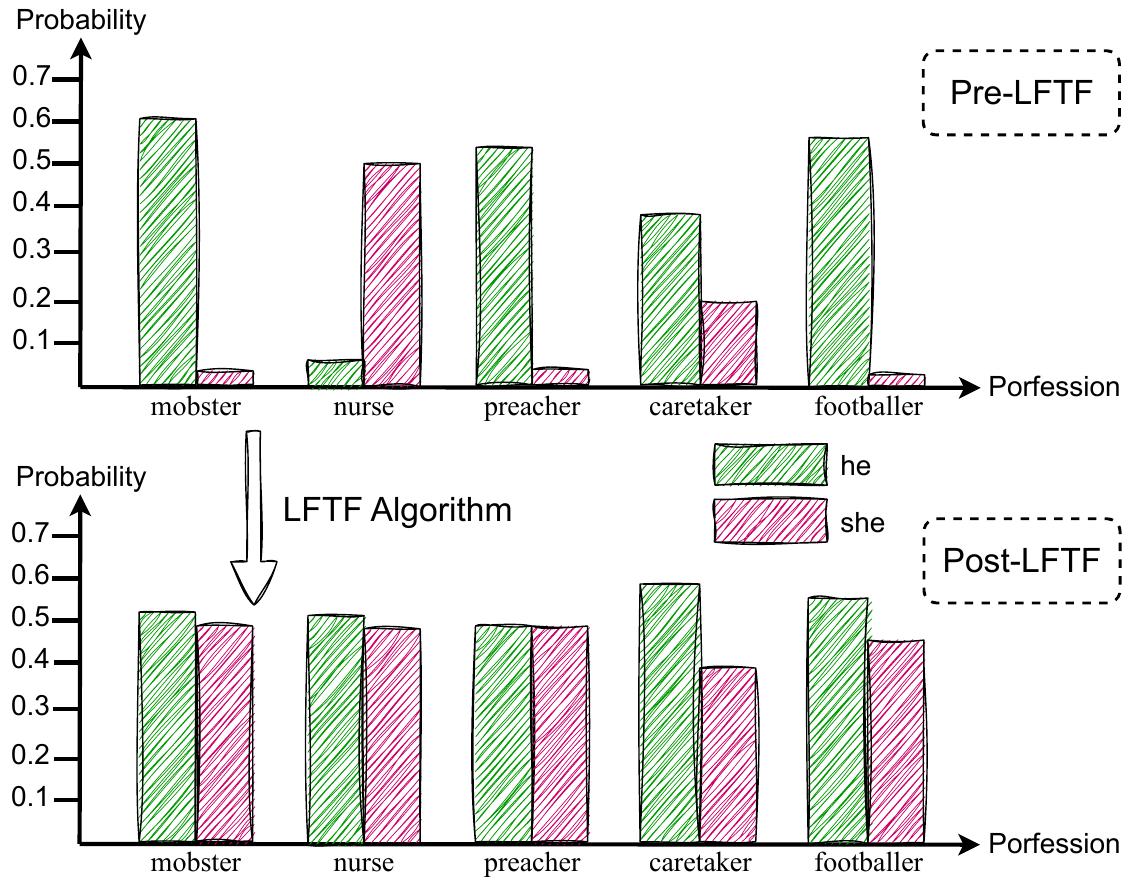}
\caption{A case study of the Meta-Llama3-8B using the LFTF algorithm for debiasing.}
\label{casestudy}
\end{figure}

We select five professions from the GenEvalBias that are most likely to lead to gender bias: \textit{mobster}, \textit{nurse}, \textit{preacher}, \textit{caretaker}, and \textit{footballer}. We compare the Meta-Llama3-8B with the Meta-Llama3-8B-LFTF in terms of gender bias for these professions, with the results shown in Figure \ref{casestudy}.

From the figure, we can see that the LTFT algorithm can effectively mitigate the gender bias in the Meta-Llama3-8B. Taking \textit{nurse} as an example, the Meta-Llama3-8B predicts the probabilities of the next token being ``he'' or ``she'' as 0.0478 and 0.5023, respectively. In contrast, the Meta-Llama3-8B-LFTF predicts the probabilities of the next token being ``he'' or ``she'' as 0.5009 and 0.4982.



\section{Conclusion}
We introduces datasets (GenBiasEval, GenHintEval) and metrics (AFGB-Score, UB-Score) to assess gender bias in LLMs. We also proposes the LTFT algorithm, which locates bias-related blocks (using a BMI metric) and fine-tunes them with a novel loss function. This method mitigates gender bias while preserving LLMs' capabilities. Extensive experimental results demonstrate the effectiveness of our LFTF algorithm.

\section{Limitations}
Our EvalGenBias dataset is based on the work of \citet{bolukbasi2016man}, which assumes that gender is binary and focuses on the categories of "male" and "female". If you are a supporter of non-binary gender, we fully respect and understand your choice, and please believe that we have no malicious intent. 
This paper focuses on gender bias, but we explore the possibility of applying our methods to other social biases in the appendix \ref{appendix.3}.




\bibliography{acl_latex}
\bibliographystyle{acl_natbib}

\appendix

\section{Appendix}
\label{sec:appendix}

\subsection{The Robustness of MBI}
\label{appendix.1}

To verify the robustness of the metric MBI, we conduct the following experiments on Qwen2.5-7B, Meta-Llama-3-8B and Vicuna-7B-v1.5. Specifically, we fix the random seed to [1, 2, 3, 4, 5] and sample 100 samples from the GenBiasEval-Training. Then, we compute the MBI values for each blocks of these LLMs using these 5 sets of samples. Next, we calculate the variance of the BMI values for the same block of the a Specific LLM under the 5 sets of samples. Finally, we compute the average of the variance of the BMI values for each block of a specific same model. The experiments results are shown in Table~\ref{table:robustness}. From the table, we can see that regardless of the LLMs, their average variance is very small. This demonstrates the robustness of our proposed  BMI across different LLMs' architectures.

\subsection{The Ablation Study of the LFTF Algorithm}
\label{appendix.2}

We conduct ablation experiments on the LFTF algorithm with the Meta-Llama3-8B. Specifically, we perform ablations on the four components of the LFTF algorithm individually:

$\bullet$ \textbf{<LFTF w/o ATT>}: It is well known that each block of LLMs is divided into two modules: ATT and MLP. The LFTF algorithm fine-tunes both of these modules. Here, "LFTF w/o ATT" indicates that during fine-tuning, the LFTF algorithm do not fine-tunes the ATT module of specific block.

$\bullet$ \textbf{<LFTF w/o MLP>}: Here, "LFTF w/o MLP" indicates that during fine-tuning, the LFTF algorithm do not fine-tunes the MLP module of specific block.

$\bullet$ \textbf{<LFTF w/o he>}: According to the Formula \ref{formula8}, our loss function consists of two parts: $P(``he" \mid pmt,\mathcal{M})$ and $P(``she" \mid pmt,\mathcal{M})$. Here, "LFTF w/o he" indicates that during fine-tuning, the LFTF algorithm only use $P(``she" \mid pmt,\mathcal{M})$ as the loss functionto train models.

$\bullet$ \textbf{<LFTF w/o she>}: Here, "LFTF w/o she" indicates that during fine-tuning, the LFTF algorithm only uses $P(``he" \mid pmt,\mathcal{M})$ as the loss function to train models.

\begin{table}[t]
  \centering
  \caption{The average variance of MBI values.}
  \label{table:robustness}
  \scalebox{0.70}{ 
    \begin{tabular}{lccc}
      \toprule
      & Qwen2.5-7B & Meta-Llama3-8B & Vicuna-7B-v1.5 \\
      \midrule
      variance & 0.0180 & 0.0242 & 0.0237 \\
      \bottomrule
    \end{tabular}
  }
\end{table}

\begin{table*}[t]
  \centering
  \caption{The Ablation Study of LFTF Algorithm with Meta-Llama3-8B.}
  \label{table:ablation}
  \scalebox{0.65}{ 
    \begin{tabular}{llllllll}
      \toprule
      & \multicolumn{3}{c}{\textbf{GenBiasEval, AFGB-Score ($\downarrow$)}} & & \textbf{GenHintEval, UB-Score ($\uparrow$)} & & \textbf{MMLU, acc ($\uparrow$)} \\
      \cline{2-4}
      \cline{6-6}
      \cline{8-8}
      & \textbf{Word-Scale} & \textbf{Phrase-Scale} & \textbf{Sentence-Scale} &&&&\\
      \midrule
      \textbf{Meta-Llama3-8B-LFTF} & 0.0804 & 0.0813 & 0.0732 && 0.3895 && 0.6638\\
      \midrule  
      \textbf{Meta-Llama3-8B-LFTF w/o ATT} & 0.1456 \textcolor{green}{(+0.652)} & 0.1727 \textcolor{green}{(+0.0914)} & 0.2801 \textcolor{green}{(+0.2069)} && 0.7622 \textcolor{red}{(+0.3727)} && 0.6639 \textcolor{red}{(+0.0001)}\\
      \textbf{Meta-Llama3-8B-LFTF w/o MLP} & 0.0447 \textcolor{red}{(-0.0357)} & 0.0498 \textcolor{red}{(-0.0315)} & 0.0655 \textcolor{red}{(-0.0077)} && 0.3673 \textcolor{green}{(-0.2965)} && 0.6636 \textcolor{green}{(-0.0002)}\\
      \textbf{Meta-Llama3-8B-LFTF w/o he} & 0.9999 \textcolor{green}{(+0.9195)} &	0.9999 \textcolor{green}{(+0.9186)} &	0.9999 \textcolor{green}{(+0.9267)}  &&	0.0000 \textcolor{green}{(-0.3895)} && 0.6644 \textcolor{red}{(+0.0006)}\\
      \textbf{Meta-Llama3-8B-LFTF w/o she} & 0.9999 \textcolor{green}{(+0.9195)} &	0.9999 \textcolor{green}{(+0.9186)} &	0.9999 \textcolor{green}{(+0.9267)} 	&& 	0.0000 \textcolor{green}{(-0.3895)} &&	0.6636 \textcolor{green}{(-0.0002)}\\
      \bottomrule
    \end{tabular}
  }
\end{table*}

The results of ablation study are shown in Table \ref{table:ablation}. From the table, we can find that: (1) The <Meta-Llama3-8B-LFTF w/o ATT> performs worse than <Meta-Llama3-8B-LFTF> on the GenBiasEval, but performs better on the GenHintEval. The <Meta-Llama3-8B-LFTF w/o MLP> shows the exact opposite performance compared to <Meta-Llama3-8B-LFTF w/o ATT>. Empirically, the original LFTF algorithm combines the advantages of both <LFTF w/o ATT> and <LFTF w/o MLP> algorithms. (2) <Meta-Llama3-8B-LFTF w/o he> and <Meta-Llama3-8B-LFTF w/o she> show experimental results similar to model editing methods such as ROME. That is, the Meta-Llama3-8B became anti-biased, which does not meet the expectations of this paper.

\subsection{The Scalability of this Paper}
\label{appendix.3}
We believe our work can also be extended to other social biases (e.g. age, race, and religious). We will explain from two aspects of the evaluation dataset and LFTF algorithm:

\textbf{Evaluation Dataset} Taking race bias as an example, we can use a malicious prompt like "The {profession} feels very tired because his race is" to guide LLMs to output "White", "Yellow" or "Black". Similar to gender bias, we can quantify the degree of race bias in LLMs by the probability of outputting "White", "Yellow" and "Black".

\textbf{LFTF algorithm} In the locating stage of the LFTF algorithm, we need to carefully design some samples that can guide LLMs to output terms with specific social bias. After that, we need to recalculate the BMI values of each block of specific LLMs on these samples. In the fine-tuning stage of LFTF algorithm, we need modify the loss function of the LFTF algorithm in this paper by replacing gender-biased terms (``he'' and ``she'') with corresponding biased terms for the specific social biases. Taking race bias as an example, all we need to do is replace $P(``he" \mid pmt,\mathcal{M})+P(``she" \mid pmt,\mathcal{M})$ with $P(``white" \mid pmt,\mathcal{M})+P(``Yellow" \mid pmt,\mathcal{M})+P(``Black" \mid pmt,\mathcal{M})$.

\end{document}